%% file: acl.tex
\pdfoutput=1

\documentclass[11pt]{article}

\usepackage[]{EMNLP2022}

\usepackage{times}
\usepackage{latexsym}

\usepackage[T1]{fontenc}

\usepackage[utf8]{inputenc}

\usepackage{soul}
\usepackage{url}
\usepackage{ulem}
\usepackage{booktabs}
\usepackage{amsmath}
\usepackage{amsthm} 
\usepackage{amsfonts}  
\usepackage{amssymb}
\usepackage{multicol}
\usepackage{multirow}
\usepackage{graphicx}
\usepackage{booktabs}
\usepackage{colortbl}
\usepackage{stfloats}
\usepackage{comment}
\usepackage{makecell}
\usepackage[noend]{algpseudocode}
\usepackage{algorithmicx,algorithm}
\usepackage{pifont}
\usepackage{color}
\usepackage{microtype}
\usepackage{cleveref}
\newcommand{\tabincell}[2]{\begin{tabular}{@{}#1@{}}#2\end{tabular}}
\title{Group is better than individual: Exploiting  Label Topologies and Label Relations for Joint Multiple Intent Detection and Slot Filling
}

\author{Bowen Xing$^{1,2}$ \and Ivor W. Tsang$^{2,1}$ \\ 
$^1$Australian Artificial Intelligence Institute, University of Technology Sydney, Australia \\
$^2$Centre for Frontier Artificial Intelligence Research, A*STAR, Singapore\\ 
\texttt{\normalsize{bwxing714@gmail.com, ivor.tsang@gmail.com}}}

\begin{document}
\maketitle
\input{abstract}
\input{introduction}

\input{method}
\input{experiment}

\input{relatedwork}
\input{conclusion}
\section*{Acknowledgements}
This work was supported by Australian Research Council  Grant DP200101328.
Bowen Xing and Ivor W. Tsang were also supported by A$^*$STAR Centre for Frontier AI Research.
\normalem
\typeout{}\bibliography{ref}
\bibliographystyle{acl_natbib}

\appendix

\section{Experiment Results on Pre-trained Language Model}
\label{sec:appendix roberta}
To explore the effect of the pre-trained language model, we use the pre-trained RoBERTa \cite{roberta} encoder\footnote{We use the RoBERTa-base checkpoint publicly that is available on https://huggingface.co/roberta-base.}  to replace the self-attentive encoder and RoBERTa is fin-tuned in the training process.
For each word, its first subword's hidden state at the last layer is taken as its word representation fed to the BiLSTMs in Sec. \ref{sec: lstm}.
We adopt AdamW \cite{weightdecay} optimizer to train the model with its default configuration.
Since our computation resources are limited, we did not conduct hyper-parameter searching for RoBERTa+ReLa-Net.
Thus we use the same hyper-parameters for RoBERTa+ReLa-Net with ReLa-Net (LSTM).

\begin{figure}[t]
 \centering
 \includegraphics[width = 0.45\textwidth]{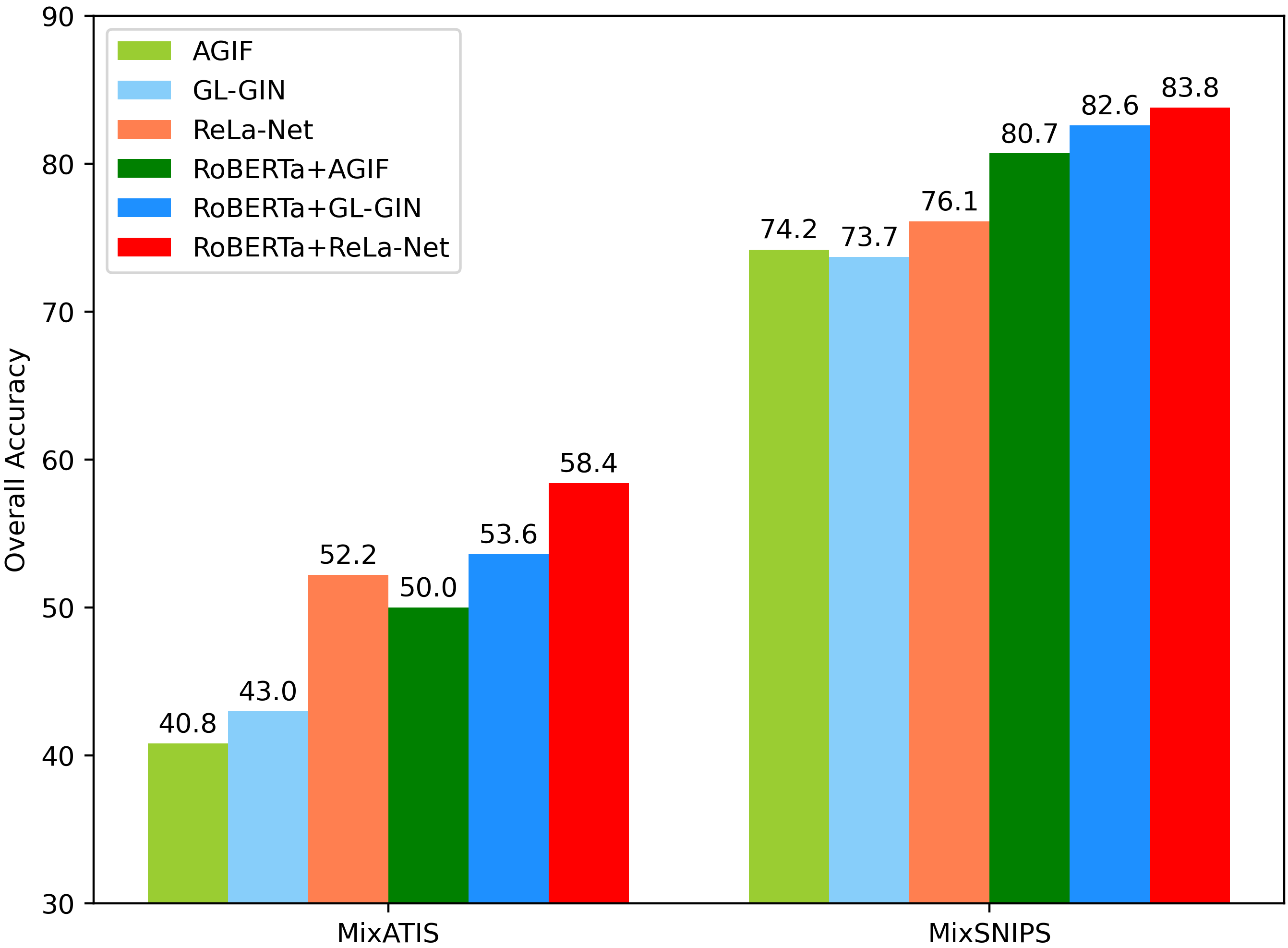}
 \caption{Overall accuracy comparison with RoBERTa.}
 \label{fig: roberta}
\end{figure}

Fig. \ref{fig: roberta} shows the results comparison of our ReLa-Net, AGIF, and GL-GIN as well as their variants using RoBERTa encoder. We have the following observations:\\
1. We can find that although the pre-trained RoBERTa encoder brings remarkable improvements via generating high-quality word representations, our RoBERTa+ReLa-Net significantly outperforms its counterparts.
This is because the core of our ReLa-Net is capturing and leveraging the correlations among the intent labels and slot labels, which is not overlapped with the advantage of the pre-trained language model focusing on semantic features. \\
2. Although RoBERTa provides much better word representations, the performances of RoBERTa-based models are still not very satisfactory, especially on MixATIS (lower than 60\%).
We think the difficulty is mainly caused by the large number of labels, e.g. there are more than 135 labels in MixATIS dataset.
Besides, our ReLa-Net (with LSTM encoder) even surpasses RoBERTa+AGIF.
This is because ReLa-Net can capture and leverage the beneficial correlations among the massive labels.
And this proves that the idea of exploiting label topologies and relations proposed in this paper is a promising perspective for improving joint multiple intent detection and slot filling.

\end{document}

%% file: abstract.tex
\begin{abstract}
Recent joint multiple intent detection and slot filling models employ label embeddings to achieve the semantics-label interactions.
However, they treat all labels and label embeddings as uncorrelated individuals, ignoring the dependencies among them. Besides, they conduct the decoding for the two tasks independently, without leveraging the correlations between them.
Therefore, in this paper, we first construct a Heterogeneous Label Graph (HLG) containing two kinds of topologies: (1) statistical dependencies based on labels' co-occurrence patterns and hierarchies in slot labels; (2) rich relations among the label nodes.
Then we propose a novel model termed ReLa-Net.
It can capture beneficial correlations among the labels from HLG.
The label correlations are leveraged to enhance semantic-label interactions. Moreover, we also propose the label-aware inter-dependent decoding mechanism to further exploit the label correlations for decoding. 
Experiment results show that our ReLa-Net significantly outperforms previous models.
Remarkably, ReLa-Net surpasses the previous best model by over 20\% in terms of overall accuracy on MixATIS dataset.
\end{abstract} 

%% file: introduction.tex
\section{Introduction}\label{sec:introduction}
Intent detection and slot filling \cite{idsf} are two fundamental tasks in spoken language understanding (SLU) \cite{slu} which is a core component in the dialog systems.
In recent years, a bunch of joint models \cite{attbirnn,slot-gated,selfgate,cmnet,sfid,qin2019,jointcap,qin2021icassp} have been proposed to tackle single intent detection and slot filling at once via modeling their correlations.

In real-world scenarios, a single utterance usually expresses multiple intents.
To handle this, multi-intent SLU \cite{kim2017} is explored and \citet{2019-joint} first propose to jointly model the multiple intent detection and slot filling in a multi-task framework.
\citet{agif} and \citet{glgin} further utilize graph attention networks (GATs) \cite{gat} which model the interactions between the embeddings of predicted intent labels and the semantic hidden states of slot filling task to leverage the dual-task correlations in an explicit way. And significantly improvements have been achieved.

\begin{figure}[t]
 \centering
 \includegraphics[width = 0.45\textwidth]{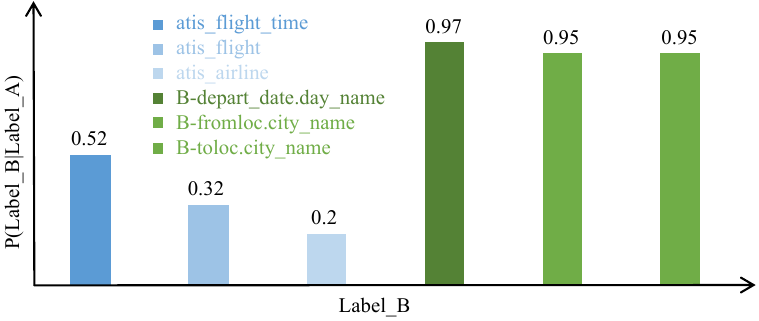}
 \caption{Illustration of top-3 high-probability occurring intent (in blue) and slot (in green) labels  when \texttt{B-depart\_date.date\_relative} (Label\_A) occurs in the training samples of MixATIS dataset.}
 \label{fig: cooccur}
\end{figure}
\begin{figure}[t]
 \centering
 \includegraphics[width = 0.43\textwidth]{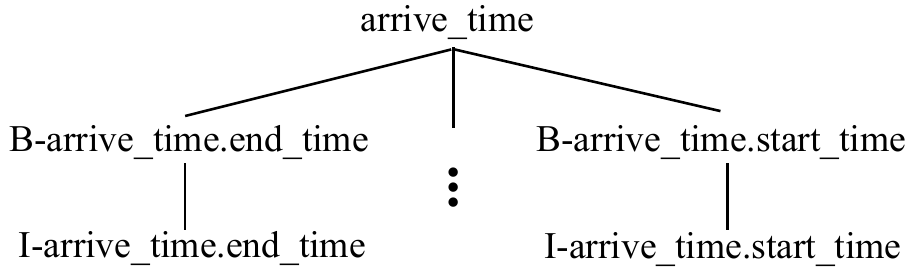}
 \caption{Illustration of the hierarchy between the pseudo label \texttt{arrive\_time} and B- slot labels, and the hierarchy between the B- slot label and its I- label.}
 \label{fig: hierarchy}
\end{figure}

However, we argue that previous works are essentially limited by ignoring the rich topological structures and relations in the joint label space of the two tasks because they treat the embeddings of intent labels and slot labels as individual parameter vectors to be learned.
Based on our observation, there are two kinds of potential topological structures in the joint label space.
(1) The global statistical dependencies among the labels based on their co-occurrence patterns, which we discover are widely-existing phenomenons.
Fig. \ref{fig: cooccur} shows an example of label co-occurance.
(2) The hierarchies in the slot labels.
Although there is no officially predefined slot label hierarchy in the datasets, we discover and define two kinds of hierarchies. An example is shown in Fig. \ref{fig: hierarchy}.
Firstly, it is intuitive that there is an intrinsic dependency between a B- slot label and its I- slot label: in the slot sequence of an utterance, an I- slot label must occur after its B- slot label, and an I- slot cannot occur solely.
Therefore, in the label space, there should be a hierarchy between a B- slot and its I- slot, while existing methods regard all slot labels as independent ones.
Secondly, we discover that some B- slot labels share the same prefix, which indicates their shared semantics.
Therefore, the prefix can be regarded as a pseudo slot label connecting these slot labels.
We believe the above statistical dependencies and hierarchies can help to capture the inner- and inter-task label correlations which benefit the joint task reasoning.
Thus in this paper, we focus on exploiting the label topologies and relations for tackling the joint task.

Besides, previous models decode the two tasks' hidden states independently without leveraging the dual-task correlations. 
This causes the misalignment of the two tasks' correct predictions.
As a result, the performance (Overall Accuracy) on sentence-level semantic frame parsing is much worse than the two tasks.
Therefore, we argue that the label embeddings conveying the dual-task correlative information should be leveraged in the decoders to guide the decoding process.

To overcome the above challenges, in this paper, we first construct a heterogeneous label graph (HLG) including both the global statistical dependencies and slot label hierarchies, based on which we define a bunch of edge types to represent the topological structures and rich relations among the labels.
Then we propose a novel model termed \textbf{R}ecurrent h\textbf{e}terogeneous \textbf{L}abel m\textbf{a}tching \textbf{Net}work (ReLa-Net) to capture the beneficial label correlative information from HLG and sufficient leverage them for tackling the joint task.
To capture the label correlative information from HLG, we design a Heterogeneous Label Graph Transformations (HLGT) module, which conducts relation-specific information aggregation among the label nodes.
We design a recurrent dual-task interacting module to leverage label correlations for semantics-label interactions.
At each step, it takes both tasks' semantic hidden states and label knowledge generated at the previous step as input.
Then sequence-reasoning BiLSTM \cite{LSTM} and GATs \cite{gat} are utilized for interactions.
As for decoding, we design a novel label-aware inter-dependent decoding mechanism that
takes both the hidden states and label embeddings as input and measures their correlation scores in the joint label embedding space.
By this means, the joint label embedding space serves as a bridge to connect the two tasks' decoding processes which then can be guided by the dual-task inter-dependencies conveyed in the learned label embeddings.
We evaluate our ReLa-Net\footnote{https://github.com/XingBowen714/ReLa-Net} on benchmark datasets and the results show that ReLa-Net achieves new state-of-the-art performance.
Further analysis proves that our method can capture nontrivial correlations among the two tasks' labels and effectively leverage them to tackle the joint task.

%% file: method.tex

\section{Problem Definition} 
Given a utterance $\mathcal{U}$, 
the task of joint multiple intent detection and slot filling aims to output a intent label set $O^I\!=\!\{o^I_1, ..., o^I_m\}$ and a slot label sequence $O^S\!=\!\{o^S_1, ..., o^S_n\}$, where $n$ is the length of $\mathcal{U}$ and $m$ is the number of intents expressed in $\mathcal{U}$.


\section{Heterogeneous Label Graph} 

\begin{figure}[t]
 \centering
 \includegraphics[width = 0.48\textwidth]{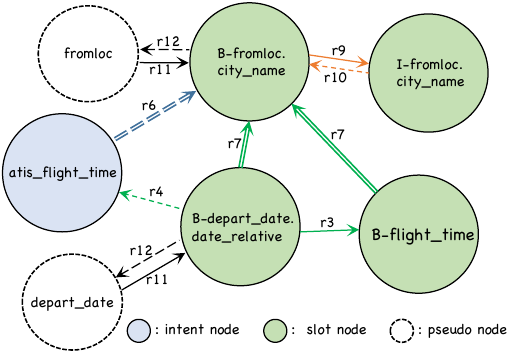}
 \caption{Illustration of a snippet from the whole HLG.}
 \label{fig: hlg}
\end{figure}

Mathematically, the HLG can be denoted as $\mathcal{G}=\left(\mathcal{V}, \mathcal{E}, \mathcal{A}, \mathcal{R}\right)$, where $\mathcal{V}$ is the set of nodes, $\mathcal{E}$ is the set of edges, $\mathcal{A}$ is the set of node types and $\mathcal{R}$ is the set of relations or edge types.
As shown in Fig. \ref{fig: hlg}, there are three types of nodes: intent nodes, slot nodes, and pseudo nodes, which correspond to the intent labels, slot labels, and pseudo slot labels. 

In HLG, there are two kinds of topologies, which correspond to statistical and hierarchical dependencies, respectively.
To represent and capture these dependencies, we define 12 relations on HLG: $r_1\sim r_8$ for statistical dependencies and $r_9\sim r_{12}$ for hierarchical dependencies, whose definitions are shown in Table \ref{table: relations}.
And $|\mathcal{V}|=N^{I} + N^{S} + N^{P}$, in which $N^{I}, N^{S}$ and $N^P$ denote the numbers of intent nodes, slot nodes and pseudo nodes. 

$r_1\sim r_8$ are based on the conditional probability between two labels, and two thresholds ($\lambda_1, \lambda_2$) are used to determine whether there is statistical dependency or strong statistical dependency from node $i$ to node $j$ regarding $P(j|i)$.
A high value of $P(j|i)$ denotes that when label $i$ occurs, there always exists label $j$ in the sample.
Namely, label $j$ can be potentially deduced when label $i$ is predicted.
In this case, an edge is established from node $i$ to node $j$ with the corresponding relation.
Note that in each sample, most slot labels are \texttt{O} which denotes the word has no meaningful slot.
Considering the slot label \texttt{O} cannot provide valuable clues, although it has high conditional probabilities (always 1.0) with other labels.
Therefore, the slot node \texttt{O} is set as an isolated node in HLG.

\begin{table}[t]
\centering
\fontsize{8}{10}\selectfont
\setlength{\tabcolsep}{1mm}{
\begin{tabular}{c|cccccccc}
\toprule
$\phi(e_{ij})$  &$\tau(i)$ & $\tau(j)$ & $P(j|i)$   \\ \midrule
$r_1$: i2i\_stat\_dep &intent &intent&$\lambda_1\leq p < \lambda_2$\\
$r_2$: i2s\_stat\_dep &intent &slot-B&$\lambda_1\leq p < \lambda_2$\\
$r_3$: s2s\_stat\_dep &slot-B &slot-B&$\lambda_1\leq p < \lambda_2$\\
$r_4$: s2i\_stat\_dep &slot-B &intent&$\lambda_1\leq p < \lambda_2$\\
$r_5$: i2i\_stat\_strong\_dep &intent &intent&$p \geq \lambda_2$\\
$r_6$: i2s\_stat\_strong\_dep &intent &slot-B&$p \geq \lambda_2$\\
$r_7$: s2s\_stat\_strong\_dep &slot-B &slot-B&$p \geq \lambda_2$\\
$r_8$: s2i\_stat\_strong\_dep &slot-B &intent&$p \geq \lambda_2$\\ \hline
$r_9$: b2i\_hierarchy &slot-B &slot-I&  $\backslash$ \\
$r_{10}$: i2b\_hierarchy &slot-I &slot-B&  $\backslash$ \\
$r_{11}$: parent2child\_hierarchy &pseudo &slot-B&  $\backslash$ \\
$r_{12}$: child2parent\_hierarchy &slot-B &pseudo&  $\backslash$ \\
\bottomrule
\end{tabular}}
\caption{Illustration of the definition of the 12 relations in HLG. $\phi(e_{ij})$ denotes the relation of edge $e_{ij}$ (from $i$ to $j$). $\tau(i)$ denotes what kinds of label the node $i$ corresponds to. $P(j|i)$ denotes the conditional probability of a sample having a label $j$ when it has a label $j$.}
\label{table: relations}
\end{table}

$r_1\sim r_8$ are based on the hierarchies we discovered in slot labels.
A B- slot node can connect and interact with many nodes, including intent nodes, other B- slot nodes, its parent pseudo node, and its I- slot node.
Differently, an I- slot node only connects with its B- slot nodes because there is a natural affiliation relation between a pair of B- and I- slot labels: an I- slot label can only appear after its B-slot label.
As for the pseudo labels we defined, there is a parent-child relationship between the parent pseudo labels and their child B- slot labels.
In HLG, a pseudo label works like an information station, which allows its children (B- slot nodes) to share their semantics and features.

\section{ReLa-Net}
\begin{figure}[t]
 \centering
 \includegraphics[width = 0.48\textwidth]{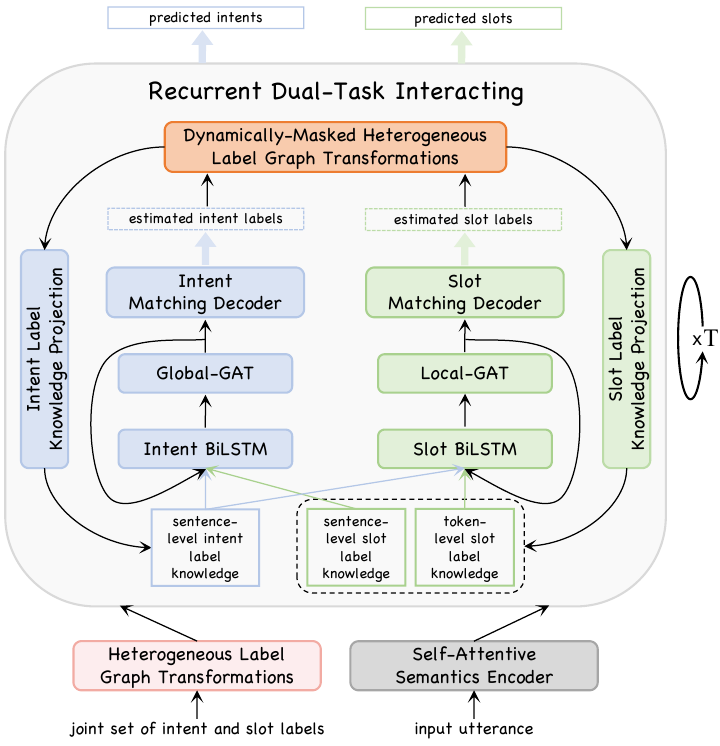}
 \caption{The network architecture of our ReLa-Net.}
 \label{fig: model}
\end{figure}

\textbf{Overview.} The architecture of our ReLa-Net is shown in Fig. \ref{fig: model}. Firstly, the initial label embeddings containing beneficial label correlations are generated by the proposed Heterogeneous Label Graph Transformations (HLGT) module, and the initial semantic hidden states are generated by the Self-Attentive Semantics Encoder.
Then we allow the two tasks to interact recurrently.
At each time step, for each task, the semantic hidden states and the label knowledge of both tasks are fused, and their correlations are learned in a BiLSTM and then a GAT.
Then the designed Label-Aware Inter-Dependent Decoding mechanism produces the estimated labels via measuring the correlations between the hidden state and the label embeddings.
The obtained labels are fed to the designed Dynamically-Masked Heterogeneous Label Graph Transformations (DM-HLGT) module, which dynamically derives the sample-specific label knowledge via operating on a DM subgraph of HLG. 
Then the obtained label knowledge is used at the next time step.
Finally, after $T$ step, the last produced labels are taken as the final predictions. 

Next, we introduce the details of each module.

\subsection{HLGT} \label{sec: label embedding}
To capture the correlations among the intent and slot labels,
inspired by \cite{rgcn,darer}, we conduct relation-specific graph transformations to achieve information aggregation on HLG:
\begin{equation}
\small
e^l_{i}=\operatorname{ReLU}(W_{1} e^{l-1}_{i} + \sum_{r \in \mathcal{R}} \sum_{j \in \mathcal{N}_{i}^{r}} \frac{1}{\left|\mathcal{N}_{i}^{r} \right|} W^r_{1} e^{l-1}_{j}) \label{eq: transformation}
\end{equation}
where $l$ denotes the layer number; $\mathcal{N}_{i}^{r}$ denotes the set of node $i$'s neighbors which are connected to it with $r$-type edges; $W_{1}$ is self weight matrix and $W^r_{1}$ is relation-specific weight matrix.
The initial node representations are obtained from an initialized matrix which is trained with the model.

After $L$ layers, the initial intent label embedding matrix $E^I$ and slot label embedding matrix $E^S$ are obtained.
Although  $E^I$ and $E^S$ contain beneficial correlative information, the HLG is globally built on the whole train set.
Therefore, they are not flexible enough for every sample, whose labels may form a local subgraph on HLG.
And we believe capturing the sample-specific label correlations can further enhance the reasoning by discovering potential neighbor labels.
To this end, we propose the Dynamically-Masked HLGT, which will be introduced in Sec. \ref{sec: dynamically-masked}.

\subsection{Self-Attentive Semantics Encoder} \label{sec: self-encoder}
Following previous works, we adopt the self-attentive encoder \cite{agif,glgin} to generate the initial semantic hidden states.
It includes a BiLSTM \cite{LSTM} and a self-attention mechanism \cite{transformer}.
The input utterance's word embeddings are fed to the BiLSTm and self-attention separately.
Then the two streams of word representations are concatenated as the output semantic hidden states.

\subsection{Recurrent Dual-Task Interacting}
\subsubsection{Semantics-Label Interaction} \label{sec: lstm}
BiLSTM has been proven to be capable of sequence reasoning \cite{seq_lstm_vaswani2016supertagging,seq_lstm_katiyar2016,seq_lstm_zheng2017joint}.
In this paper, we utilize two BiLSTMs for intent and slot to achieve the fusion and interactions among the semantics and label knowledge of both tasks. 
Specifically, the two BiLSTMs can be formulated as:
\begin{equation}
 \small 
 \begin{aligned}
 \hat{h}^{I,i}_t &= \operatorname{BiLSTM}_{\text{I}}(h^{I,i}_t\|K^I_t\|K^S_t, \hat{h}^{I,i-1}_t, \hat{h}^{I,i+1}_t)\\
\hat{h}^{S,i}_t &= \operatorname{BiLSTM}_{\text{S}}(h^{S,i}_t\|K^I_t\|K^{S,i}_t, \hat{h}^{S,i-1}_t, \hat{h}^{S,i+1}_t)
\end{aligned}
\end{equation}
where $\|$ denotes concatenation operation.

For Intent BiLSTM ($\operatorname{BiLSTM}_{\text{I}}$), the input is the concatenation of intent semantic hidden states $h^{I,i}_t$, sentence-level intent label knowledge $K^I_t$ and the sentence-level slot label knowledge $K^S_t$, where $t$ denotes the step number of recurrent dual-task interacting.
Here we use sentence-level label knowledge because multiple intent detection is on sentence-level.
For Slot BiLSTM ($\operatorname{BiLSTM}_{\text{S}}$), the input is the concatenation of slot semantic hidden states $h^{S,i}_t$, sentence-level intent label knowledge $K^I_t$ and the token-level slot label knowledge $K^{S,i}_t$.
Here we use token-level slot label knowledge because slot filling is a token-level task.
And we use sentence-level intent label knowledge here because it can provide indicative clues of potential slot labels regarding the captured inter-task dependencies among intent and slot labels.

At the first step, there is no available label knowledge, so the inputs of the two BiLSTMs are the initial  hidden states generated in Sec. \ref{sec: self-encoder}.
How to obtain $K^I_t, K^{S}_t, K^{S,i}_t$ is depicted in Sec. \ref{sec: label knowledge}.

\subsubsection{Graph Attention Networks}
\paragraph{Global-GAT}
Since multiple intent detection is a sentence-level task, we believe it is beneficial to further capture the global dependencies among the words.
We first construct a fully-connected graph on the input utterance.
There are $n$ nodes in this graph and each one corresponds to a word.
And we adopt the GAT \cite{gat} for information aggregation.
The initial representation of node $i$ is $\hat{h}^{I,i}_t$ generated by $\operatorname{BiLSTM}_{\text{I}}$.
After $L$ layers, we obtain $\widetilde{H}^I_t = \{\widetilde{h}^{I,i}_t\}_{i=1}^N$.

\paragraph{Local-GAT} Since slot filling is on token-level, we adopt a Local-GAT to capture the token-level local dependencies.
The Local-GAT works on a locally-connected graph where node $i$ is connected to nodes $\{i-w, ..., i+w\}$, where sliding window size $w$ is a hyper-parameter.
The initial representation of node $i$ is $\hat{h}^{S,i}_t$ generated by $\operatorname{BiLSTM}_{\text{S}}$.
After $L$ layers, we obtain $\widetilde{H}^S_t = \{\widetilde{h}^{S,i}_t\}_{i=1}^N$.
And we adopt the GAT \cite{gat} for information aggregation.
\subsubsection{Label-Aware Inter-Dependent Decoding}
\begin{figure}[t]
 \centering
\setlength{\abovecaptionskip}{4pt}
 \includegraphics[width = 0.45\textwidth]{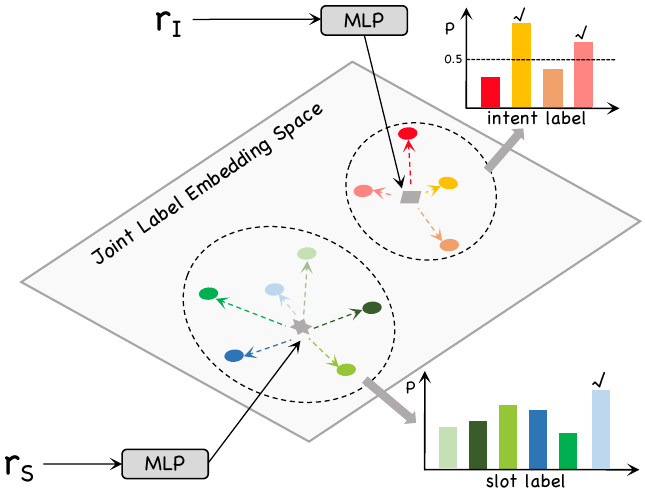}
\caption{Illustration of our proposed label-aware co-decoding mechanism. Blue/green circles denote slot labels and red/yellow circles denote intent labels. The dash arrow denotes the distance between the hidden state projection and the label. The shorter the distance, the greater the probability that the label is correct. \checkmark denotes that the label is selected as the prediction. }
 \label{fig: co-decoder}
\end{figure}

In this work, we design a novel label-aware inter-dependent decoding mechanism (shown in Fig. \ref{fig: co-decoder}) to leverage the dual-task correlative information conveyed in the learned label embeddings to guide the decoding process.
Concretely, the hidden state is first projected into the joint label embedding space, and then the dot products are conducted between it and all label embeddings of a specific task to obtain the correlation score vector. 
The larger correlation score indicates the shorter distance between the hidden state's projection and the label embedding. Thus the hidden state more likely belongs to the corresponding class.
In this way, the joint label embedding space serves as a bridge that connects the decoding processes of the two tasks, and the beneficial correlative information conveyed in the learned label embeddings can guide the inter-dependent decoding for the two tasks.
Next, we introduce the intent and slot decoders in detail.

\paragraph{Intent Decoder}
Following previous works, we conduct token-level multi-label classification. 
Firstly, the intent hidden state $\widetilde{h}^{I,i}_{t}$ is projected into the joint label embedding space via an MLP.
Then the correlation score vector $C_t^{I,i}$ is calculated via dot products between it and all intent label embeddings.
This process can be formulated as:
\begin{equation}
\small
C_t^{I,i} = ({W}_{I}(\operatorname{LeakyReLU}({W}_{h}^I \widetilde{h}^{I,i}_{t}+{b}_{h}))+{b}_{I})\ \widehat{E}_t^{I} \label{eq: decoding}
\end{equation}
where $C_t^{I,i}$ is a $N^I$-dimensional vector; $\widehat{E}_t^{I}$ is the sample-specific intent label embedding matrix at step $t$; $W_*$ and $b_*$ are model parameters.

Then $C_t^{I,i}$ is fed to the sigmoid function to obtain the token-level intent probability vector.
And a threshold (0.5) is used to select the intents.
Finally, we obtain the predicted sentence-level intents by voting for all tokens' intents \cite{glgin}.

\paragraph{Slot Decoder}
Following \cite{glgin}, we conduct parallel decoding.
Similar to Eq. \ref{eq: decoding}, the slot correlation score vector $C_t^{S,i}$ is obtained by:
\begin{equation}
\small
C_t^{S,i} = ({W}_{S}(\operatorname{LeakyReLU}({W}_{h}^S \widetilde{h}^{S,i}_{t}+{b}_{S}))+{b}_{S})\ \widehat{E}_t^{S}
\end{equation}
where $C_t^{S,i}$ is a $N^S$-dimensional vector; $\widehat{E}_t^{S}$ is the sample-specific slot label embedding matrix at step $t$; $W_*$ and $b_*$ are model parameters.

Then $C_t^{I,i}$ is fed to the softmax function to obtain the slot probability distribution.
Finally, the argmax function is used to select the predicted slot.

\subsubsection{DM-HLGT} \label{sec: dynamically-masked}
After decoding, we obtain the predicted intents and slots at step $t$.
To capture the sample-specific label correlations, which we believe further benefit the reasoning in the next step, we first construct a DM subgraph of HLG for each sample.
The process is simple: the nodes of predicted intent labels and slot labels and their first-order neighbors are reserved, while other nodes on HLG are masked out.
Then relation-specific graph transformations are conducted on this graph to achieve information aggregation, whose formulation is similar to Eq. \ref{eq: transformation}.
At the first step, the $\widehat{E}_0^{I}$ and $\widehat{E}_0^{S}$ are ${E}^{I}$ and ${E}^{S}$.

\subsubsection{Label Knowledge Projection} \label{sec: label knowledge}
To provide label knowledge for next time step, the predicted labels should be projected into vectors.
In Sec. \ref{sec: lstm}, we use three kinds of label knowledge: $K^I_t, K^{S}_t, K^{S,i}_t$.
$K^I_t$ and $K^{S}_t$ are sentence-level intent and slot label knowledge, respectively.
$K^I_t$ is the sum of label embeddings of the predicted intents and $K^{S}_t$ is the sum of label embeddings of the predicted slots while the \texttt{O} slot is excluded.
$K^{S,i}_t$ is token-level slot label knowledge, which is the label embedding of the predicted slot of token $t$.
\subsection{Optimization}
Following previous works, the standard loss function for intent task ($\mathcal{L}_{I}$) and slot task ($\mathcal{L}_{S}$) are:
\begin{equation}
 \small 
 \begin{split}
l(\hat{y}, y)&=\hat{y} \log (y)+(1-\hat{y}) \log (1-y)\\
\mathcal{L}_{I} &=\sum_{t=1}^T\sum_{i=1}^{n} \sum_{j=1}^{N^{I}} l(\hat{y}_{i}^{I}[j], y_{i}^{I,t}[j])\\
\mathcal{L}_{S} &= \sum_{t=1}^T\sum_{i=1}^{n} \sum_{j=1}^{N^{S}} \hat{y}_{i}^{S}[j] \log (y_{i}^{[S,t]}[j])
\end{split}
\end{equation}
Besides, we design a constraint loss ($\mathcal{L}_{cst}$) to push ReLa-Net to generate better label distributions at step $t$ than step $t-1$:
\begin{equation}
\small
\mathcal{L}_{cons}^{I} \!=\! \sum_{t=2}^T \sum_{i=1}^n \sum_{j=1}^{{N}^I} \hat{y}_{i}^{I}[j] \ \operatorname{max} (0, y_i^{[I,t-1]}[j]\!-\!y_i^{[I,t]}[j])
\end{equation}
And $\mathcal{L}^S_{cst}$ can be derived similarly.

Then the final training objective is:
\begin{equation}
 \small 
\mathcal{L} = \gamma_I\left(\mathcal{L}_I + \beta_I \mathcal{L}_{cst}^{I}\right) + \gamma_S\left(\mathcal{L}_S + \beta_S \mathcal{L}_{cst}^{S}\right)
\end{equation}
where $\gamma_I=0.1$ and $\gamma_S=0.9$ balance the two tasks; $\beta_I=0.01$ and $\beta_S=1.0$ balance the standard loss and constraint loss for the two tasks.

%% file: experiment.tex
\section{Experiments}
\subsection{Settings}
\textbf{Datasets.}
Following previous works, we conduct experiments on two benchmarks: MixATIS and MixSNIPS  \cite{atis,snips,agif}.
In MixATIS, the split of train/dev/test set is 13162/756/828 (utterances).
In MixSNIPS, the split of train/dev/test set is 39776/2198/2199 (utterances).\\
\textbf{Evaluation Metrics.}
Following previous works, we evaluate multiple intent detection using accuracy (Acc), slot filling using F1 score, and sentence-level semantic frame parsing using overall Acc.
Overall Acc denotes the ratio of utterances for which both intents and slots are predicted correctly.\\
\textbf{Implementation Details.}
Following previous works, the word and label embeddings are randomly initialized and trained with the model. Due to limited space, the experiments using pre-trained language model is presented in the Appendix.
The dimension of the word/label embedding is 128 on MixATIS and 256 on MixSNIPS.
For HLG, $\lambda_1=0.4$ and $\lambda_2=0.9$.
The hidden dim is 200.
The max layer number of all GNNs is 2.
Max time step $T$ is 2.
We adopt Adam \cite{adam} optimizer to train our model using the default setting.
For all experiments, we select the best model on the dev set and report its results on the test set.
Our source code will be released.

\subsection{Main Results}

We compare our model with: (1) Attention BiRNN \cite{attbirnn}; (2) Slot-Gated \cite{slot-gated}; (3) Bi-Model \cite{bimodel}; (4) SF-ID Network \cite{sfid}; (5) Stack-Propagation \cite{qin2019}; (6) Joint Multiple ID-SF \cite{idsf}; (7) AGIF \cite{agif}; (8) GL-GIN \cite{glgin}.
Table \ref{table: results} lists the results on the test sets.
We can observe that:

\noindent 1. Our Rela-Net consistently outperforms all baselines by large margins on all datasets and tasks.
Specifically, compared with Gl-GIN, the previous best model, ReLa-Net achieves an absolute improvement of 9.2\% in terms of Overall Acc on MixATIS, a relative improvement of over 20\%.\\
2. ReLa-Net gains larger improvements on the intent task than slot task.
The reason is that ReLa-Net is the first model to allow the two tasks to interact with each other, while previous models only leverage the predicted intents to guide slot filling.\\
3. Generally, the results on overall Acc are obviously worse than slot F1 and intent Acc, indicating that it is hard to align the correct prediction of intents and slots.
Compared with baselines, our ReLa-Net achieves especially significant improvements on overall Acc and effectively reduces the performance gap between overall Acc and slot F1/intent Acc.
This can be attributed to the fact that our model can capture sufficient and beneficial label correlations from HLG, and the designed label-aware inter-dependent decoding mechanism can leverage the label correlations to guide decoding, making the two tasks' decoding processes correlative. 
As a result, the correct predictions of the two tasks are better aligned. \\
4. ReLa-Net gains more significant improvements on MixATIS dataset than MixSNIPS.
The reason is that MixATIS dataset includes much more labels than MixSNIPS: MixATIS includes 17 intent labels and 117 slot labels,  while MixSNIPS only include 7 intent labels and 72 slot labels.
More labels in MixATIS dataset cause it is much harder for correct predictions, which can be proved by the lower Overall Acc scores of all models on MixATIS than MixSNIPS.
However, the core strength of our model is leveraging the label topologies and relations.
Therefore, more labels in MixATIS dataset result in larger improvements. 

\begin{table}[t]
\centering
\fontsize{8}{10}\selectfont
\setlength{\tabcolsep}{0.9mm}{
\begin{tabular}{l|ccc|ccc}
\toprule
\multirow{2}{*}{Models} & \multicolumn{3}{c|}{MixATIS} & \multicolumn{3}{c}{MixSNIPS} \\ \cline{2-7} 
                        & \tabincell{c}{Overall\\ (Acc) } &\tabincell{c}{Slot \\(F1)}  &\tabincell{c}{Intent\\(Acc)}& \tabincell{c}{Overall\\ (Acc) } &\tabincell{c}{Slot \\(F1)}  &\tabincell{c}{Intent\\(Acc)}           \\ \midrule
Attention BiRNN  &  39.1 &  86.4      &   74.6      & 59.5        &  89.4     & 95.4 \\
Slot-Gated     &  35.5 &  87.7      &   63.9      & 55.4        &  87.9     & 94.6 \\
Bi-Model          &  34.4 &  83.9      &   70.3      & 63.4        &  90.7    & 95.6 \\
SF-ID                &  34.9 &  87.4      &   66.2      & 59.9        &  90.6     & 95.0 \\
Stack-Propagation &  40.1 &  87.8      &   72.1      & 72.9        &  94.2     & 96.0 \\
Joint Multiple ID-SF &36.1 &84.6    &   73.4      & 62.9        &  90.6     & 95.1 \\
AGIF                 &  40.8 &  86.7      &   74.4      & \underline{74.2}        &  \underline{94.2}     & 95.1 \\
GL-GIN &  \underline{43.0} &  \underline{88.2}      &   \underline{76.3}      & 73.7        &  94.0     & \underline{95.7} \\ \midrule
ReLa-Net (ours)           &  \textbf{52.2} &\textbf{90.1} &\textbf{78.5} & \textbf{76.1}  &  \textbf{94.7} & \textbf{97.6} \\
  \bottomrule
\end{tabular}}
\caption{Main results. ReLa-Net obtains statistically significant improvements over baselines with $p<0.01$.} 
\label{table: results}
\vspace{-4mm}
\end{table}

\subsection{Ablation Study}

\begin{table}[b]
\centering
\fontsize{8}{10}\selectfont
\setlength{\tabcolsep}{2mm}{
\begin{tabular}{l|ccc}
\hline
\multirow{2}{*}{Variants} & \multicolumn{3}{c}{MixATIS} \\ \cline{2-4} 
                        & \tabincell{c}{Overall\\ (Acc) } &\tabincell{c}{Slot \\(F1)}  &\tabincell{c}{Intent\\(Acc)}   \\ \hline
ReLa-Net &\textbf{52.2} &\textbf{90.1} &\textbf{78.5}  \\ \hline
w/o stat\_dep &45.3 ($\downarrow$6.9) &{87.5}($\downarrow$2.6) &{76.9}($\downarrow$1.6)  \\
w/o hierarchy &48.4 ($\downarrow$3.8)  &{87.9}($\downarrow$2.2) &{78.1}($\downarrow$0.4)  \\
w/o relation &45.1($\downarrow$7.1) &{88.2}($\downarrow$1.9) &{75.4}($\downarrow$3.1)  \\
\hline
w/o matching &{45.2}($\downarrow$7.0) &{88.1}($\downarrow$2.0) &{77.4}($\downarrow$1.1)  \\
w/o GATs &{47.6}($\downarrow$4.6) &{88.3}($\downarrow$1.8) &{77.2}($\downarrow$1.3)  \\
w/o DM-HLGT &{50.2}($\downarrow$2.0) &{89.2}($\downarrow$0.9) &{78.0}($\downarrow$0.5)  \\
  \hline
\end{tabular}}
\caption{Results of ablation experiments.} 
\label{table: ablation}
\end{table}
We conduct two groups of ablation experiments on HLG and ReLa-Net to verify the necessities of their components.
Table \ref{table: ablation} shows the results.

\noindent \textbf{HLG.} 
\textit{\textbf{w/o stat\_dep}} denotes the statistical dependencies are removed, which means there are no edges between intent and slot label nodes on HLG.
We can observe sharp drops in the performances of two tasks and the semantic frame parsing.
This proves that statistical dependencies can effectively capture beneficial inter-task dependencies. 
\textit{\textbf{w/o hierarchy}} denotes the slot hierarchies are removed.
We can find although Intent Acc is not seriously affected, the Slot F1 and Overall Acc are both significantly reduced.
This proves that slot hierarchies can effectively bring improvements via capturing the beneficial structural dependencies among slot labels.
\textit{\textbf{w/o relation}} denotes HLG collapses into a homogeneous graph without specific relations.
Its poor performances prove that the defined rich relations on HLG are crucial for capturing the label correlations, which play key roles in dual-task reasoning and label-aware inter-dependent decoding.
\begin{figure}[t]
 \centering
 \includegraphics[width = 0.45\textwidth]{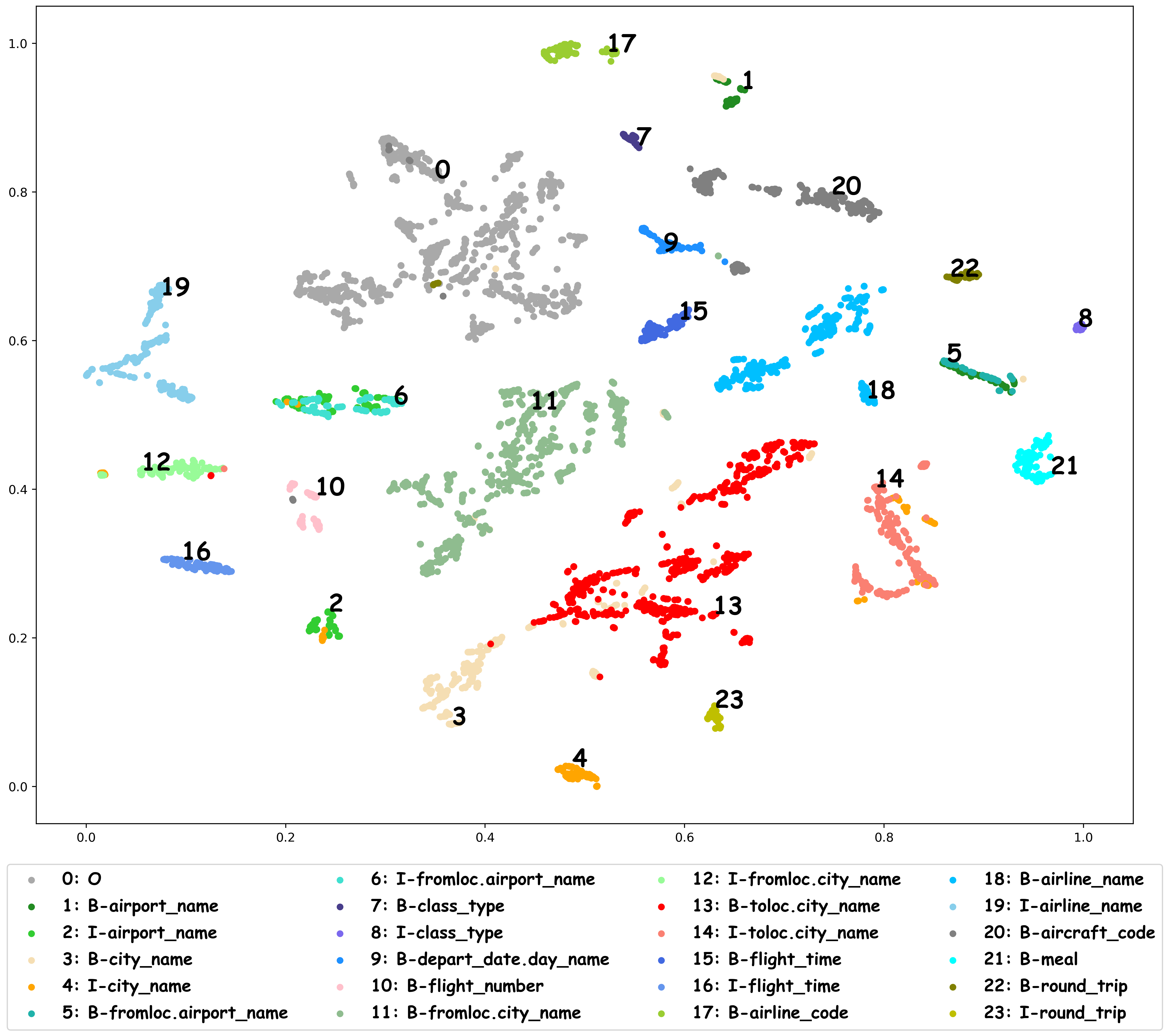}
 \caption{Visualization of ReLa-Net's slot clusters. }
 \label{fig: tsne_slot}
\end{figure}
\begin{figure}[t]
 \centering
 \includegraphics[width = 0.45\textwidth]{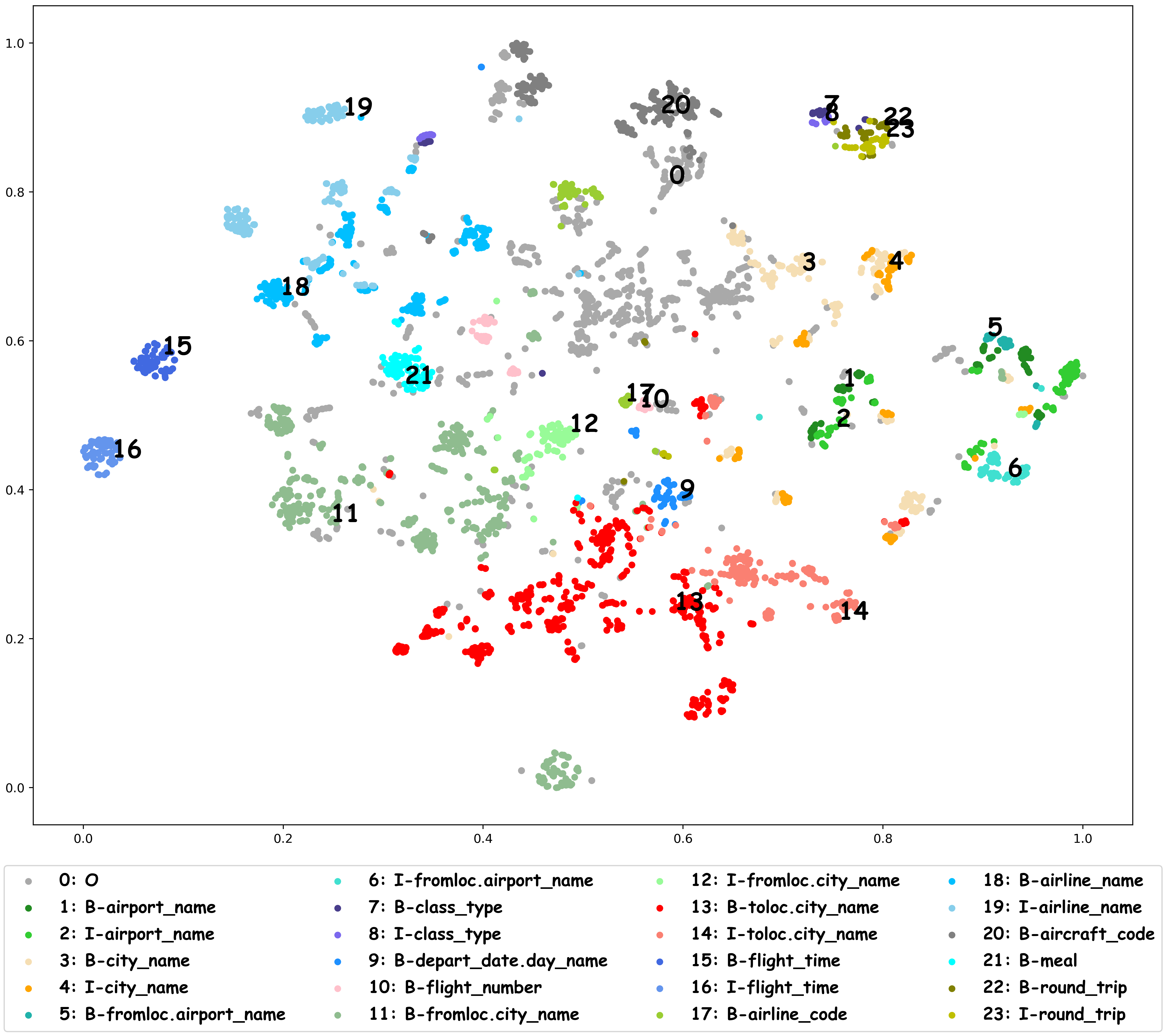}
 \caption{Visualization of GL-GIN's slot clusters. }
 \label{fig: tsne_slot_glgin}
\end{figure}
\noindent\textbf{ReLa-Net.}
\textit{\textbf{w/o matching}} denotes the designed matching decoders in ReLa-Net are replaced with the linear decoders used in previous models.
Its worse results prove that our designed matching decoders can effectively improve the decoding quality via leveraging the beneficial label correlations captured from HLG.
\textit{\textbf{w/o GATs}} denotes the Global-GAT and Local-GAT are removed.
Its worse results prove that the sentence-level global dependencies and the token-level local dependencies are important for the intent task and slot task, respectively.
However, even if its semantics and semantics-label interactions are only modeled via LSTM, it still outperforms all baselines.
This can be attributed to the fact that ReLa-Net can learn beneficial label correlations from HLG and sufficiently leverage them for decoding.
\textit{\textbf{w/o DM-HLGT}} denotes the DM-HGT is removed.
And its results verify that DM-HLGT can effectively enhance the performance via capturing the sample-specific label correlations and discovering potential neighbor labels.

\subsection{Visualization of Hidden State Clusters}

To better understand the promising results of ReLa-Net, we perform TSNE \cite{tsne} visualization on the final hidden states of the utterances words in the test set of MixATIS.
It is hard to visualize intent states because a single hidden state usually corresponds to multiple intents.
Therefore, we visualize the final slot hidden state of each word in the utterances.
Since there are more than 100 slot labels, to simplify the visualization, we rank the slot labels regarding their frequencies in the test set, and we show the results of the top-24 labels.
Besides, most slot labels in the test set are \texttt{O}, and we randomly pick 500 ones.
The slot clusters visualization of our ReLa-Net is shown in Fig. \ref{fig: tsne_slot}.


From Fig. \ref{fig: tsne_slot}, we can observe that the boundaries of the clusters are clear, and the data points in the same cluster are tightly closed to each other.
And we can find that the B- slot clusters and their corresponding I- slot clusters are clearly separated. 
Besides, there is nearly no incorrect I- slot data point.
The high accuracy of I- slot labels prove that our ReLa-Net effectively resolves the uncoordinated slot problem (e.g., \texttt{B-singer} followed by \texttt{I-song}).
This is attributed to the beneficial slot hierarchies we discovered and leveraged.

For comparison, we visualize GL-GIN’s slot clusters in the same way, as shown in Fig. \ref{fig: tsne_slot_glgin}.
We can observe that GL-GIN's slot clusters boundaries are not clear.
Some data points of `\texttt{O}' slot are close to other clusters.
And some B- clusters and their I- clusters even significantly overlap each other.

The above observations prove that our method can learn better hidden states via effectively capturing and leveraging the label correlations.
\subsection{Visualization of Label Correlations}
\begin{figure*}[t]
 \centering
 \includegraphics[width =\textwidth]{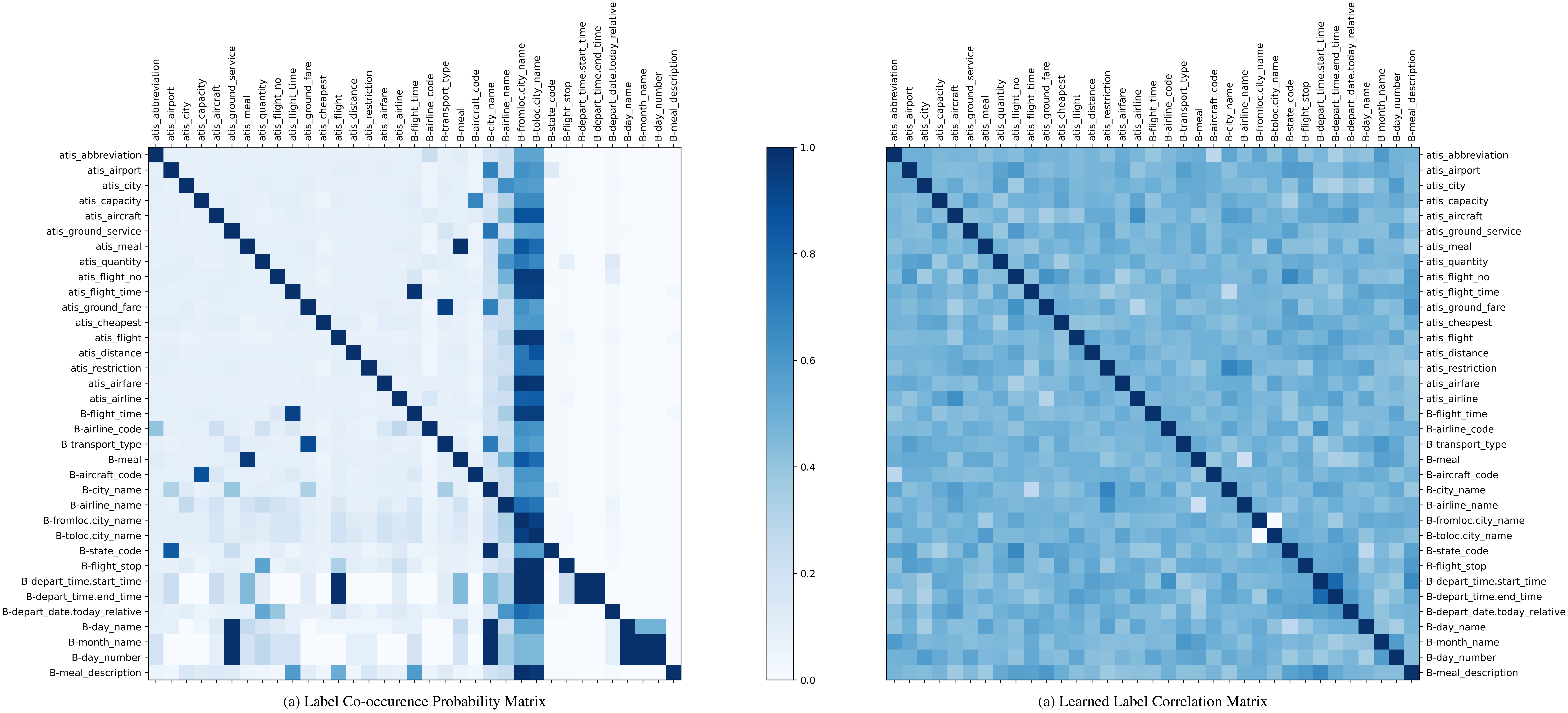}
 \caption{Visualization of label co-occurrence matrix and learned label correlation matrix. }
 \label{fig: label}
\end{figure*}

To further look into the label correlations, we visualize the label co-occurrence probability matrix constructed from MixATIS training set and the correlation matrix of ReLa-Net's learned label embeddings, as shown in Fig. \ref{fig: label} (a) and (b).
Label correlations of two labels are measured by the cosine similarity of their embeddings. For simplification, we visualize all intent labels (17 ones) and some slot labels (18 ones).
From Fig.\ref{fig: label} (a), we can observe that there exist many distinct co-occurrence patterns among the labels of the two tasks.
The core of this work is capturing and leveraging the beneficial label correlations, and promising improvements have been observed from the experiment results, while previous works neglect this point.
From Fig.\ref{fig: label} (b), we can find that our ReLa-Net learns non-trivial label embeddings.
From the patterns of label correlations in Fig. \ref{fig: label} (b), we can hardly observe the label co-occurrence patterns in Fig. \ref{fig: label} (a).
This proves that ReLa-Net can comprehensively leverage the label co-occurrences to learn robust and effective label embeddings, rather than just mechanically memorize them.

%% file: relatedwork.tex
\section{Related Work}\label{sec: relatedwork}
To capture and leverage the dual-task correlations, a group of models \cite{ijcai2016joint,hakkani2016multi,slot-gated,selfgate,sfid,cmnet,qin2019,jointcap,slotrefine,qin2021icassp,ni2021recent} have been proposed for joint intent detection and slot filling.
However, they only consider the single-intent scenario.
Unlike them, we tackle the joint task of multiple intent detection and slot filling, which is more challenging and practical in the real-world scenario.

Recently, multi-intent SLU has received increasing attention since it can handle complex utterances expressing multiple intents.
\citet{kim2017} propose a two-stage model to tackle the multiple intent detection task.
And \citet{2019-joint} propose the first joint model that employs a multi-task framework including a slot-gate mechanism to tackle the joint task of multiple intent detection and slot filling.
With the wide utilization of graph neural networks (GNNs) in various NLP tasks \cite{kagnet,HGNN4MHQA,qin2021co,dignet,neuralsubgraph,jair,tetci}, state-of-the-art multi-intent SLU systems also leverage GNNs to model the cross-task interactions.
\citet{agif} utilize GATs to introduce the fine-grained information of the predicted multiple intents into slot filling in an adaptive way.
Further, \citet{glgin} propose a GAT-based model in which slot filling is conducted in a non-autoregressive manner.

Previous models ignore the correlations among the two tasks' labels, treating their embeddings as separated parameters to be learned.
Our method is significantly different from previous ones. This is the first work to discover, capture and leverage the topologies and relations among the labels for joint multiple intent detection and slot filling.

%% file: conclusion.tex
\section{Conclusion}\label{sec: conclusion}
In this work, we improve joint multiple intent detection and slot filling from a new perspective: exploiting label typologies and relations.
Specifically, we first design a heterogeneous label graph to represent the statistical dependencies and hierarchies in rich relations.
Then we propose a recurrent heterogeneous label matching network to end-to-end capture and leverage the beneficial label correlations.
Experiments demonstrate that our method significantly outperforms previous models, achieving new state-of-the-art.
Further studies verify the advantages of our methods and show that our model can generate better hidden states and learns non-trivial label embeddings.
In the future, we will apply our method in other multi-task scenarios.

\section*{Limitations}

The core of this work is proving the power of the learned label embeddings obtained by exploiting label topologies and relations.
Therefore, to better demonstrate that the significant improvements come from the learned label embeddings, we do not focus on enhancing the interactions between the label knowledge and semantics. We adopt BiLSTM, a relatively basic module, to model the interactions between the prediction information and semantics.
However, there is no doubt that BiLSTM is not the optimal module for these label-semantics interactions, and in practice, many modules can be applied to achieve this better.
For instance, GL-GIN adopts not only BiLSTM but the proposed global-locally graph interaction layers working on the designed label-semantic graph to model the label-semantics interactions.
Therefore, although our ReLa-Net significantly outperforms state-of-the-art models, intuitively, its performance is limited by the simple module for modeling the label-semantics interactions.
In our future work, we will design more advanced modules for semantics-label interactions.

%% file: acl.bbl
\begin{thebibliography}{40}
\expandafter\ifx\csname natexlab\endcsname\relax\def\natexlab#1{#1}\fi

\bibitem[{Coucke et~al.(2018)Coucke, Saade, Ball, Bluche, Caulier, Leroy,
  Doumouro, Gisselbrecht, Caltagirone, Lavril, Primet, and Dureau}]{snips}
Alice Coucke, Alaa Saade, Adrien Ball, Théodore Bluche, Alexandre Caulier,
  David Leroy, Clément Doumouro, Thibault Gisselbrecht, Francesco Caltagirone,
  Thibaut Lavril, Maël Primet, and Joseph Dureau. 2018.
\newblock \href {http://arxiv.org/abs/1805.10190} {Snips voice platform: an
  embedded spoken language understanding system for private-by-design voice
  interfaces}.

\bibitem[{E et~al.(2019)E, Niu, Chen, and Song}]{sfid}
Haihong E, Peiqing Niu, Zhongfu Chen, and Meina Song. 2019.
\newblock \href {https://doi.org/10.18653/v1/P19-1544} {A novel bi-directional
  interrelated model for joint intent detection and slot filling}.
\newblock In \emph{Proceedings of the 57th Annual Meeting of the Association
  for Computational Linguistics}, pages 5467--5471, Florence, Italy.
  Association for Computational Linguistics.

\bibitem[{Fang et~al.(2020)Fang, Sun, Gan, Pillai, Wang, and Liu}]{HGNN4MHQA}
Yuwei Fang, Siqi Sun, Zhe Gan, Rohit Pillai, Shuohang Wang, and Jingjing Liu.
  2020.
\newblock \href {https://aclanthology.org/2020.emnlp-main.710} {Hierarchical
  graph network for multi-hop question answering}.
\newblock In \emph{Proceedings of the 2020 Conference on Empirical Methods in
  Natural Language Processing (EMNLP)}, pages 8823--8838, Online. Association
  for Computational Linguistics.

\bibitem[{Gangadharaiah and Narayanaswamy(2019)}]{2019-joint}
Rashmi Gangadharaiah and Balakrishnan Narayanaswamy. 2019.
\newblock \href {https://doi.org/10.18653/v1/N19-1055} {Joint multiple intent
  detection and slot labeling for goal-oriented dialog}.
\newblock In \emph{Proceedings of the 2019 Conference of the North {A}merican
  Chapter of the Association for Computational Linguistics: Human Language
  Technologies, Volume 1 (Long and Short Papers)}, pages 564--569, Minneapolis,
  Minnesota. Association for Computational Linguistics.

\bibitem[{Goo et~al.(2018)Goo, Gao, Hsu, Huo, Chen, Hsu, and Chen}]{slot-gated}
Chih-Wen Goo, Guang Gao, Yun-Kai Hsu, Chih-Li Huo, Tsung-Chieh Chen, Keng-Wei
  Hsu, and Yun-Nung Chen. 2018.
\newblock \href {https://doi.org/10.18653/v1/N18-2118} {Slot-gated modeling for
  joint slot filling and intent prediction}.
\newblock In \emph{Proceedings of the 2018 Conference of the North {A}merican
  Chapter of the Association for Computational Linguistics: Human Language
  Technologies, Volume 2 (Short Papers)}, pages 753--757, New Orleans,
  Louisiana. Association for Computational Linguistics.

\bibitem[{Hakkani-Tür et~al.(2016)Hakkani-Tür, Tur, Celikyilmaz, Chen, Gao,
  Deng, and Wang}]{hakkani2016multi}
Dilek Hakkani-Tür, Gokhan Tur, Asli Celikyilmaz, Yun-Nung Chen, Jianfeng Gao,
  Li~Deng, and Ye-Yi Wang. 2016.
\newblock \href {https://doi.org/10.21437/Interspeech.2016-402} {Multi-domain
  joint semantic frame parsing using bi-directional rnn-lstm}.
\newblock In \emph{Interspeech 2016}, pages 715--719.

\bibitem[{Hemphill et~al.(1990)Hemphill, Godfrey, and Doddington}]{atis}
Charles~T. Hemphill, John~J. Godfrey, and George~R. Doddington. 1990.
\newblock \href {https://aclanthology.org/H90-1021} {The {ATIS} spoken language
  systems pilot corpus}.
\newblock In \emph{Speech and Natural Language: Proceedings of a Workshop Held
  at Hidden Valley, {P}ennsylvania, June 24-27,1990}.

\bibitem[{Hochreiter and Schmidhuber(1997)}]{LSTM}
Sepp Hochreiter and Jürgen Schmidhuber. 1997.
\newblock \href {https://doi.org/10.1162/neco.1997.9.8.1735} {Long short-term
  memory}.
\newblock \emph{Neural Computation}, 9(8):1735--1780.

\bibitem[{Katiyar and Cardie(2016)}]{seq_lstm_katiyar2016}
Arzoo Katiyar and Claire Cardie. 2016.
\newblock \href {https://doi.org/10.18653/v1/P16-1087} {Investigating {LSTM}s
  for joint extraction of opinion entities and relations}.
\newblock In \emph{Proceedings of the 54th Annual Meeting of the Association
  for Computational Linguistics (Volume 1: Long Papers)}, pages 919--929,
  Berlin, Germany. Association for Computational Linguistics.

\bibitem[{Kim et~al.(2017)Kim, Ryu, and Lee}]{kim2017}
Byeongchang Kim, Seonghan Ryu, and Gary~Geunbae Lee. 2017.
\newblock \href {https://doi.org/10.1007/s11042-016-3724-4} {Two-stage
  multi-intent detection for spoken language understanding}.
\newblock \emph{Multimedia Tools and Applications}, 76(9):11377--11390.

\bibitem[{Kingma and Ba(2015)}]{adam}
Diederik~P. Kingma and Jimmy Ba. 2015.
\newblock \href {http://arxiv.org/abs/1412.6980} {Adam: {A} method for
  stochastic optimization}.
\newblock In \emph{3rd International Conference on Learning Representations,
  {ICLR} 2015, San Diego, CA, USA, May 7-9, 2015, Conference Track
  Proceedings}.

\bibitem[{Li et~al.(2018)Li, Li, and Qi}]{selfgate}
Changliang Li, Liang Li, and Ji~Qi. 2018.
\newblock \href {https://doi.org/10.18653/v1/D18-1417} {A self-attentive model
  with gate mechanism for spoken language understanding}.
\newblock In \emph{Proceedings of the 2018 Conference on Empirical Methods in
  Natural Language Processing}, pages 3824--3833, Brussels, Belgium.
  Association for Computational Linguistics.

\bibitem[{Lin et~al.(2019)Lin, Chen, Chen, and Ren}]{kagnet}
Bill~Yuchen Lin, Xinyue Chen, Jamin Chen, and Xiang Ren. 2019.
\newblock \href {https://doi.org/10.18653/v1/D19-1282} {{K}ag{N}et:
  Knowledge-aware graph networks for commonsense reasoning}.
\newblock In \emph{Proceedings of the 2019 Conference on Empirical Methods in
  Natural Language Processing and the 9th International Joint Conference on
  Natural Language Processing (EMNLP-IJCNLP)}, pages 2829--2839, Hong Kong,
  China. Association for Computational Linguistics.

\bibitem[{Liu and Lane(2016)}]{attbirnn}
Bing Liu and Ian Lane. 2016.
\newblock \href {https://doi.org/10.21437/Interspeech.2016-1352}
  {{Attention-Based Recurrent Neural Network Models for Joint Intent Detection
  and Slot Filling}}.
\newblock In \emph{Proc. Interspeech 2016}, pages 685--689.

\bibitem[{Liu et~al.(2019{\natexlab{a}})Liu, Meng, Zhang, Zhou, Chen, and
  Xu}]{cmnet}
Yijin Liu, Fandong Meng, Jinchao Zhang, Jie Zhou, Yufeng Chen, and Jinan Xu.
  2019{\natexlab{a}}.
\newblock \href {https://doi.org/10.18653/v1/D19-1097} {{CM}-net: A novel
  collaborative memory network for spoken language understanding}.
\newblock In \emph{Proceedings of the 2019 Conference on Empirical Methods in
  Natural Language Processing and the 9th International Joint Conference on
  Natural Language Processing (EMNLP-IJCNLP)}, pages 1051--1060, Hong Kong,
  China. Association for Computational Linguistics.

\bibitem[{Liu et~al.(2019{\natexlab{b}})Liu, Ott, Goyal, Du, Joshi, Chen, Levy,
  Lewis, Zettlemoyer, and Stoyanov}]{roberta}
Yinhan Liu, Myle Ott, Naman Goyal, Jingfei Du, Mandar Joshi, Danqi Chen, Omer
  Levy, Mike Lewis, Luke Zettlemoyer, and Veselin Stoyanov. 2019{\natexlab{b}}.
\newblock Roberta: A robustly optimized bert pretraining approach.
\newblock \emph{arXiv preprint arXiv:1907.11692}.

\bibitem[{Loshchilov and Hutter(2019)}]{weightdecay}
Ilya Loshchilov and Frank Hutter. 2019.
\newblock \href {https://openreview.net/forum?id=Bkg6RiCqY7} {Decoupled weight
  decay regularization}.
\newblock In \emph{ICLR}.

\bibitem[{Ni et~al.(2021)Ni, Young, Pandelea, Xue, Adiga, and
  Cambria}]{ni2021recent}
Jinjie Ni, Tom Young, Vlad Pandelea, Fuzhao Xue, Vinay Adiga, and Erik Cambria.
  2021.
\newblock \href {https://arxiv.org/abs/2105.04387} {Recent advances in deep
  learning based dialogue systems: A systematic survey}.
\newblock \emph{arXiv preprint arXiv:2105.04387}.

\bibitem[{Qin et~al.(2019)Qin, Che, Li, Wen, and Liu}]{qin2019}
Libo Qin, Wanxiang Che, Yangming Li, Haoyang Wen, and Ting Liu. 2019.
\newblock \href {https://doi.org/10.18653/v1/D19-1214} {A stack-propagation
  framework with token-level intent detection for spoken language
  understanding}.
\newblock In \emph{Proceedings of the 2019 Conference on Empirical Methods in
  Natural Language Processing and the 9th International Joint Conference on
  Natural Language Processing (EMNLP-IJCNLP)}, pages 2078--2087, Hong Kong,
  China. Association for Computational Linguistics.

\bibitem[{Qin et~al.(2021{\natexlab{a}})Qin, Li, Che, Ni, and Liu}]{qin2021co}
Libo Qin, Zhouyang Li, Wanxiang Che, Minheng Ni, and Ting Liu.
  2021{\natexlab{a}}.
\newblock Co-gat: A co-interactive graph attention network for joint dialog act
  recognition and sentiment classification.
\newblock In \emph{Proceedings of the AAAI Conference on Artificial
  Intelligence}, volume~35, pages 13709--13717.

\bibitem[{Qin et~al.(2021{\natexlab{b}})Qin, Liu, Che, Kang, Zhao, and
  Liu}]{qin2021icassp}
Libo Qin, Tailu Liu, Wanxiang Che, Bingbing Kang, Sendong Zhao, and Ting Liu.
  2021{\natexlab{b}}.
\newblock \href {https://doi.org/10.1109/ICASSP39728.2021.9414110} {A
  co-interactive transformer for joint slot filling and intent detection}.
\newblock In \emph{ICASSP 2021 - 2021 IEEE International Conference on
  Acoustics, Speech and Signal Processing (ICASSP)}, pages 8193--8197.

\bibitem[{Qin et~al.(2021{\natexlab{c}})Qin, Wei, Xie, Xu, Che, and
  Liu}]{glgin}
Libo Qin, Fuxuan Wei, Tianbao Xie, Xiao Xu, Wanxiang Che, and Ting Liu.
  2021{\natexlab{c}}.
\newblock \href {https://doi.org/10.18653/v1/2021.acl-long.15} {{GL}-{GIN}:
  Fast and accurate non-autoregressive model for joint multiple intent
  detection and slot filling}.
\newblock In \emph{Proceedings of the 59th Annual Meeting of the Association
  for Computational Linguistics and the 11th International Joint Conference on
  Natural Language Processing (Volume 1: Long Papers)}, pages 178--188, Online.
  Association for Computational Linguistics.

\bibitem[{Qin et~al.(2020)Qin, Xu, Che, and Liu}]{agif}
Libo Qin, Xiao Xu, Wanxiang Che, and Ting Liu. 2020.
\newblock \href {https://doi.org/10.18653/v1/2020.findings-emnlp.163} {{AGIF}:
  An adaptive graph-interactive framework for joint multiple intent detection
  and slot filling}.
\newblock In \emph{Findings of the Association for Computational Linguistics:
  EMNLP 2020}, pages 1807--1816, Online. Association for Computational
  Linguistics.

\bibitem[{Schlichtkrull et~al.(2018)Schlichtkrull, Kipf, Bloem, van~den Berg,
  Titov, and Welling}]{rgcn}
Michael~Sejr Schlichtkrull, Thomas~N. Kipf, Peter Bloem, Rianne van~den Berg,
  Ivan Titov, and Max Welling. 2018.
\newblock \href {https://doi.org/10.1007/978-3-319-93417-4_38} {Modeling
  relational data with graph convolutional networks}.
\newblock In \emph{ESWC}, pages 593--607.

\bibitem[{Tur and De~Mori(2011)}]{idsf}
Gokhan Tur and Renato De~Mori. 2011.
\newblock \emph{Spoken language understanding: Systems for extracting semantic
  information from speech}.
\newblock John Wiley \& Sons.

\bibitem[{van~der Maaten and Hinton(2008)}]{tsne}
Laurens van~der Maaten and Geoffrey Hinton. 2008.
\newblock \href {http://jmlr.org/papers/v9/vandermaaten08a.html} {Visualizing
  data using t-sne}.
\newblock \emph{Journal of Machine Learning Research}, 9(86):2579--2605.

\bibitem[{Vaswani et~al.(2016)Vaswani, Bisk, Sagae, and
  Musa}]{seq_lstm_vaswani2016supertagging}
Ashish Vaswani, Yonatan Bisk, Kenji Sagae, and Ryan Musa. 2016.
\newblock \href {https://doi.org/10.18653/v1/N16-1027} {Supertagging with
  {LSTM}s}.
\newblock In \emph{Proceedings of the 2016 Conference of the North {A}merican
  Chapter of the Association for Computational Linguistics: Human Language
  Technologies}, pages 232--237, San Diego, California. Association for
  Computational Linguistics.

\bibitem[{Vaswani et~al.(2017)Vaswani, Shazeer, Parmar, Uszkoreit, Jones,
  Gomez, Kaiser, and Polosukhin}]{transformer}
Ashish Vaswani, Noam Shazeer, Niki Parmar, Jakob Uszkoreit, Llion Jones,
  Aidan~N. Gomez, Lukasz Kaiser, and Illia Polosukhin. 2017.
\newblock \href
  {https://proceedings.neurips.cc/paper/2017/hash/3f5ee243547dee91fbd053c1c4a845aa-Abstract.html}
  {Attention is all you need}.
\newblock In \emph{Advances in Neural Information Processing Systems 30: Annual
  Conference on Neural Information Processing Systems 2017, December 4-9, 2017,
  Long Beach, CA, {USA}}, pages 5998--6008.

\bibitem[{Velickovic et~al.(2018)Velickovic, Cucurull, Casanova, Romero,
  Li{\`{o}}, and Bengio}]{gat}
Petar Velickovic, Guillem Cucurull, Arantxa Casanova, Adriana Romero, Pietro
  Li{\`{o}}, and Yoshua Bengio. 2018.
\newblock \href {https://openreview.net/forum?id=rJXMpikCZ} {Graph attention
  networks}.
\newblock In \emph{6th International Conference on Learning Representations,
  {ICLR} 2018, Vancouver, BC, Canada, April 30 - May 3, 2018, Conference Track
  Proceedings}. OpenReview.net.

\bibitem[{Wang et~al.(2018)Wang, Shen, and Jin}]{bimodel}
Yu~Wang, Yilin Shen, and Hongxia Jin. 2018.
\newblock \href {https://doi.org/10.18653/v1/N18-2050} {A bi-model based {RNN}
  semantic frame parsing model for intent detection and slot filling}.
\newblock In \emph{Proceedings of the 2018 Conference of the North {A}merican
  Chapter of the Association for Computational Linguistics: Human Language
  Technologies, Volume 2 (Short Papers)}, pages 309--314, New Orleans,
  Louisiana. Association for Computational Linguistics.

\bibitem[{Wu et~al.(2020)Wu, Ding, Lu, and Xie}]{slotrefine}
Di~Wu, Liang Ding, Fan Lu, and Jian Xie. 2020.
\newblock \href {https://doi.org/10.18653/v1/2020.emnlp-main.152}
  {{S}lot{R}efine: A fast non-autoregressive model for joint intent detection
  and slot filling}.
\newblock In \emph{Proceedings of the 2020 Conference on Empirical Methods in
  Natural Language Processing (EMNLP)}, pages 1932--1937, Online. Association
  for Computational Linguistics.

\bibitem[{Xing and Tsang(2022{\natexlab{a}})}]{darer}
Bowen Xing and Ivor Tsang. 2022{\natexlab{a}}.
\newblock \href {https://aclanthology.org/2022.findings-acl.286} {{DARER}:
  Dual-task temporal relational recurrent reasoning network for joint dialog
  sentiment classification and act recognition}.
\newblock In \emph{Findings of the Association for Computational Linguistics:
  ACL 2022}, pages 3611--3621, Dublin, Ireland. Association for Computational
  Linguistics.

\bibitem[{Xing and Tsang(2022{\natexlab{b}})}]{dignet}
Bowen Xing and Ivor Tsang. 2022{\natexlab{b}}.
\newblock Dignet: Digging clues from local-global interactive graph for
  aspect-level sentiment classification.
\newblock \emph{arXiv preprint arXiv:2201.00989}.

\bibitem[{Xing and Tsang(2022{\natexlab{c}})}]{neuralsubgraph}
Bowen Xing and Ivor Tsang. 2022{\natexlab{c}}.
\newblock \href {https://doi.org/10.24963/ijcai.2022/614} {Neural subgraph
  explorer: Reducing noisy information via target-oriented syntax graph
  pruning}.
\newblock In \emph{Proceedings of the Thirty-First International Joint
  Conference on Artificial Intelligence, {IJCAI-22}}, pages 4425--4431.

\bibitem[{Xing and Tsang(2022{\natexlab{d}})}]{jair}
Bowen Xing and Ivor~W Tsang. 2022{\natexlab{d}}.
\newblock Out of context: A new clue for context modeling of aspect-based
  sentiment analysis.
\newblock \emph{Journal of Artificial Intelligence Research}, 74:627--659.

\bibitem[{Xing and Tsang(2022{\natexlab{e}})}]{tetci}
Bowen Xing and Ivor~W. Tsang. 2022{\natexlab{e}}.
\newblock \href {https://doi.org/10.1109/TETCI.2022.3156989} {Understand me, if
  you refer to aspect knowledge: Knowledge-aware gated recurrent memory
  network}.
\newblock \emph{IEEE Transactions on Emerging Topics in Computational
  Intelligence}, 6(5):1092--1102.

\bibitem[{Young et~al.(2013)Young, Gašić, Thomson, and Williams}]{slu}
Steve Young, Milica Gašić, Blaise Thomson, and Jason~D. Williams. 2013.
\newblock \href {https://doi.org/10.1109/JPROC.2012.2225812} {Pomdp-based
  statistical spoken dialog systems: A review}.
\newblock \emph{Proceedings of the IEEE}, 101(5):1160--1179.

\bibitem[{Zhang et~al.(2019)Zhang, Li, Du, Fan, and Yu}]{jointcap}
Chenwei Zhang, Yaliang Li, Nan Du, Wei Fan, and Philip Yu. 2019.
\newblock \href {https://doi.org/10.18653/v1/P19-1519} {Joint slot filling and
  intent detection via capsule neural networks}.
\newblock In \emph{Proceedings of the 57th Annual Meeting of the Association
  for Computational Linguistics}, pages 5259--5267, Florence, Italy.
  Association for Computational Linguistics.

\bibitem[{Zhang and Wang(2016)}]{ijcai2016joint}
Xiaodong Zhang and Houfeng Wang. 2016.
\newblock \href {http://www.ijcai.org/Abstract/16/425} {A joint model of intent
  determination and slot filling for spoken language understanding}.
\newblock In \emph{Proceedings of the Twenty-Fifth International Joint
  Conference on Artificial Intelligence, {IJCAI} 2016, New York, NY, USA, 9-15
  July 2016}, pages 2993--2999. {IJCAI/AAAI} Press.

\bibitem[{Zheng et~al.(2017)Zheng, Wang, Bao, Hao, Zhou, and
  Xu}]{seq_lstm_zheng2017joint}
Suncong Zheng, Feng Wang, Hongyun Bao, Yuexing Hao, Peng Zhou, and Bo~Xu. 2017.
\newblock \href {https://doi.org/10.18653/v1/P17-1113} {Joint extraction of
  entities and relations based on a novel tagging scheme}.
\newblock In \emph{Proceedings of the 55th Annual Meeting of the Association
  for Computational Linguistics (Volume 1: Long Papers)}, pages 1227--1236,
  Vancouver, Canada. Association for Computational Linguistics.

\end{thebibliography}
